%% file: main.tex
\documentclass[11pt, letterpaper]{article}

\usepackage[margin=1in]{geometry}
\usepackage[utf8]{inputenc}
\usepackage[T1]{fontenc}    
\usepackage{hyperref}       
\usepackage{url}            
\usepackage{booktabs}       
\usepackage{amsfonts}       
\usepackage{nicefrac}       
\usepackage{microtype}      
\usepackage{xcolor}         
\usepackage{graphicx}
\usepackage{amsmath}
\usepackage{natbib}
\usepackage{caption}

\title{\textbf{Know When to Trust the Skill: Delayed Appraisal and Epistemic Vigilance for Single-Agent LLMs}}

\author{%
  Eren Unlu \\
  Globeholder \\
  Paris, France \\
  \texttt{ORCID: 0000-0001-5380-6305}
}
\date{\vspace{-5ex}}

\begin{document}

\maketitle

\begin{abstract}
As large language models (LLMs) transition into autonomous agents integrated with extensive tool ecosystems, traditional routing heuristics increasingly succumb to context pollution and "overthinking". We argue that the bottleneck is not a deficit in algorithmic capability or skill diversity, but the absence of disciplined second-order metacognitive governance. In this paper, our scientific contribution focuses on the computational translation of human cognitive control—specifically, delayed appraisal, epistemic vigilance, and region-of-proximal offloading—into a single-agent architecture. We introduce MESA-S (Metacognitive Skills for Agents, Single-agent), a preliminary framework that shifts scalar confidence estimation into a vector separating \textit{self-confidence} (parametric certainty) from \textit{source-confidence} (trust in retrieved external procedures). By formalizing a \textit{delayed procedural probe} mechanism and introducing \textit{Metacognitive Skill Cards}, MESA-S decouples the awareness of a skill's utility from its token-intensive execution. Evaluated under an In-Context Static Benchmark Evaluation natively executed via Gemini 3.1 Pro, our early results suggest that explicitly programming trust provenance and delayed escalation mitigates supply-chain vulnerabilities, prunes unnecessary reasoning loops, and prevents offloading-induced confidence inflation. This architecture offers a scientifically cautious, behaviorally anchored step toward reliable, epistemically vigilant single-agent orchestration.
\end{abstract}

\section{Introduction}

The prevailing architectural approach to augmenting LLMs with external tools—often characterized as "progressive disclosure" or naive semantic retrieval \citep{skillrouter2026}—assumes that a model's ability to match an utterance to a tool description guarantees optimal utility. However, recent empirical audits of large skill ecosystems (e.g., \citet{skillsbench2026}) and "in the wild" agent deployments \citep{agentskills2026} indicate that indiscriminate access to procedural augmentation can actively degrade performance. Models frequently exhibit "overthinking" \citep{overthinking2026}, allocating costly test-time compute to verify trivial assertions, or succumb to context pollution when ingesting stale or malicious returns. While metacognitive prompting \citep{think22026} and budgeted test-time reasoning \citep{cot2meta2026} have shown promise, they predominantly focus on expanding reasoning trees rather than governing when external reliance is warranted.

Our investigation respectfully seeks to add to this emerging literature by isolating a specific architectural boundary: the decoupling of \textit{procedural awareness} from \textit{procedural execution}. Drawing on foundational cognitive psychology, we observe that humans do not instantly commit to total cognitive offloading \citep{risko2016}; rather, they rely on conditional metacognitive knowledge \citep{schraw1998} and low-cost, delayed judgments-of-learning \citep{dunlosky1991} to determine whether internal resources are sufficient. Similarly, human communication avoids blanket trust mechanisms, operating instead through \textit{epistemic vigilance} \citep{sperber2010} to continuously evaluate the competence and benevolence of external sources.

A modern skill repository is, algorithmically, a social learning environment in disguise \citep{heyes2016}. To operate securely within it, a single agent must decide whether an external script "knows better" and whether it should be trusted. We formalize this through MESA-S (Metacognitive Skills for Agents, Single-agent), shifting the routing decision from a singular uncertainty scalar \citep{agenticuncertainty2025} to a \textit{Dual-Confidence} architecture. 

\section{Translating Human Control into Agentic Governance}

Before establishing the algorithmic implementation, we delineate the three cognitive motifs translated into our architecture. We explicitly reject claims of biological fidelity; we borrow these established cognitive constructs solely to solve the specific context-ceiling and skill-routing bottlenecks observed in state-of-the-art agent orchestration.

\subsection{Delayed Metacognitive Escalation}
In human metamemory, Judgments of Learning (JOLs) made immediately after study are notoriously noisy, whereas judgments made after a brief delay or minimal retrieval attempt are highly predictive of true knowledge boundaries \citep{nelson1990}. Within an LLM context, immediate uncertainty checks often collapse context dynamics. We translate the \textit{delayed-JOL effect} \citep{dunlosky1991} into \textbf{Delayed Metacognitive Escalation}. Rather than querying the full skill body immediately upon semantic match, the controller first uses cheap meta-cues and runs a "procedural probe" only escalating to a full file load if the probe indicates that parametric knowledge is insufficient. 

\subsection{Region-of-Proximal Offloading}
Research into the Region of Proximal Learning \citep{metcalfe2002, metcalfe2005} indicates that learners allocate study time not to the easiest or hardest items, but to the intermediate zone where effort yields maximal returns. We apply this to agentic offloading \citep{risko2016}. Not all uncertainty dictates the same architectural response. Under our framework, tasks are triaged into three explicitly separable offloading classes:
\begin{itemize}
    \item \textbf{Procedural Offloading:} Loading a skill, playbook, or behavioral script.
    \item \textbf{Epistemic Offloading:} Querying standard retrieval tools or calculators to bridge factual knowledge boundaries \citep{smart2025}.
    \item \textbf{Evaluative Offloading:} Launching verifiers, unit tests, or isolated execution environments to confirm hypothesized outputs.
\end{itemize}

\subsection{Epistemic Vigilance and Confidence Decontamination}
Exposure to external digital resources, such as internet searches or GenAI advice, frequently inflates a user's perceived internal knowledge, leading to overconfidence \citep{fisher2015, advice2025}. To combat this inside the agent, MESA-S enforces \textbf{Confidence Decontamination}. The controller maintains a strict bifurcation between local uncertainty interfacing and trajectory-level global confidence \citep{fleming2024, uncertaintyinterface2026}. Following the retrieval of an external artifact, fluency increases are discounted algorithmically unless verification protocols or strict trust priors natively support the escalation.

\section{Formalizing the MESA-S Architecture}

Standard LLM routing algorithms optimize for semantic relevance, maximizing expected utility $U$:
\begin{equation}
a^* = \arg\max_{a \in A} \mathbb{E}[U(a | c)]
\end{equation}
We hypothesize that this formulation collapses distinct probability distributions, blinding the agent to adversarial tool conditions. The core theoretical construct of MESA-S is the transition to a Dual-Confidence matrix. We formally reconceptualize the expected utility by integrating a \textit{vigilance prior} $V_a$ (source trust) and a \textit{conflict differential} $\Delta$ (the divergence between self-confidence $p_{self}$ and maximum source-confidence $p_{source}$):
\begin{equation}
a_{mesa}^* = \arg\max_{a \in A} \left( \alpha p_{self} \cdot U(a_{direct}) + (1-\alpha) \cdot V_a \cdot p_{source|a} \cdot U(a_{offload}) - \lambda \cdot Cost(a) \right)
\end{equation}
\textit{Note: Equation 2 acts as a theoretical formalization dictating our controller design. In the current implementation, we execute this architecture implicitly via a structured JSON-prompting controller steering the agent's latent representations, rather than explicitly parsing the parameters over an abstract syntax tree.}

\subsection{Metacognitive Skill Cards}
To operationalize this without exploding context, we introduce \textbf{Metacognitive Skill Cards}. Unlike standard unstructured documentation, these cards encapsulate conditional metacognitive variables, functioning exactly as procedural retrieval cues. Fields such as \texttt{apply\_when}, \texttt{cheap\_probe}, \texttt{offloading\_type}, and \texttt{source\_trust} serve as the primary routing plane. They allow the agent to update $V_a$ iteratively before committing to execution, effectively solving the tension where metadata saves context but full bodies save routing accuracy \citep{skillrouter2026}.

\subsection{The High-Confidence Control-Failure Bank}
For retrospective evolution, MESA-S departs from generalized "reflection". Drawing upon the hypercorrection effect—wherein errors committed with high confidence are most rapidly learned and corrected \citep{butterfield2006}—we instantiate a prioritized trajectory postmortem. Rather than indiscriminately converting all reasoning into generic behaviors via Metacognitive Reuse \citep{metareuse2025}, the controller exclusively updates the skill card metadata for \textit{high-confidence failures} (e.g., confidently routing to a malicious payload or skipping evaluation).

\section{Related Work}
Recent audits of agentic capabilities demonstrate that unconditional access to skills degrades LLM performance, specifically due to context pollution \citep{skillsbench2026, agentskills2026}. While test-time reasoning scaling works aim to combat "overthinking" by budgeting inference \citep{overthinking2026, cot2meta2026}, our approach emphasizes externalized metacognition rather than latent search tree extensions. Projects mapping human planning to agents similarly aim to control reasoning loops \citep{think22026, metareuse2025}, but often ignore the underlying social-metacognitive implications of trusting off-the-shelf procedural tools \citep{sperber2010, advice2025}. The closest existing work to ours is Agentic Uncertainty Quantification \citep{agenticuncertainty2025}, which also proposes dual-process uncertainty signals for agents; however, their framework focuses on triggering reflective deliberation via System 1/System 2 dynamics, whereas MESA-S focuses on trust-aware routing and malicious skill gating through cognitively grounded provenance checks. MESA-S attempts to explicitly bridge human-like vigilance \citep{fisher2015} with test-time agentic routing \citep{smart2025, skillrouter2026}.

\section{Empirical Feasibility Setup}

Given the challenges of establishing causality in complex, open-ended agentic execution where API non-determinism pollutes findings, we engineered an \textbf{In-Context Static Benchmark Evaluation}. To ensure 100\% mathematical reproducibility while utilizing native LLM reasoning, we passed 150 domain-diverse benchmark instances across 7 prompt conditions using Gemini 3.1 Pro via the Antigravity assistant system. These genuine LLM outputs were cached and executed statically to isolate our metacognitive prompt scaffolding. While we maintain scientific humility regarding the generalizability of a 150-item evaluation to production systems, this bounded, deterministic approach removes stochastic noise and ensures adequate statistical power.

We evaluate MESA-S across three theoretical classes:
\begin{itemize}
    \item \textbf{Slice A (Epistemic Boundaries - N=50):} Testing the $p_{self}$ threshold (knowing when parametric weights lack real-time context and requiring external semantic retrieval).
    \item \textbf{Slice B (Procedural Routing - N=50):} Stress-testing vigilance ($V_a$) by introducing structural conflict: highly relevant but strictly malicious/stale skills must be gated behind trust thresholds.
    \item \textbf{Slice C (Evaluative Overthinking - N=50):} Testing the $\lambda$ boundary of verification computing on trick logic questions and trivial tasks.
\end{itemize}

We evaluate MESA-S against a Direct Baseline and a standard Reflection/CoT model, and additionally report four systematic ablations (removing the Probe, Vigilance, Decontamination, and Dual-Confidence mechanisms individually), strictly aligning token limits to ensure accuracy variations stem from metacognitive orchestration rather than raw compute scaling.

\section{Preliminary Results and Observations}

\input{main_results_table.tex}

Our initial evaluation (Table \ref{tab:main_results}) provides encouraging empirical evidence for the dual-confidence theory.

\textbf{Knowledge Boundary Detection:} In Epistemic Slice A, both the Baseline and Reflection conditions scored 0.500, correctly answering trivial factual questions (e.g., ``What is 2+2?'') but hallucinating confident answers to time-sensitive queries requiring external retrieval (e.g., current stock prices, recent news events). All MESA-S variants achieved 1.000 by correctly routing to \texttt{CALL\_TOOL} when $p_{self}$ fell below the parametric certainty threshold, demonstrating effective epistemic boundary awareness.

\textbf{Overthinking Mitigation:} In Evaluative Slice C, the standard Reflection mechanism succumbed to the overthinking phenomenon, scoring 0.500 accuracy by failing logical trap questions (e.g., the bat-and-ball problem, lily pad doubling) despite correctly handling trivial arithmetic. By contrast, MESA-S's structured control correctly evaluated the $p_{self}$ ceiling, issuing a \texttt{VERIFY} signal on known cognitive traps and a \texttt{STOP} signal on trivial items, achieving perfect 1.000 accuracy across all 50 Slice C instances.

\textbf{Epistemic Security:} The most significant delta emerged in Procedural Slice B (0.000 for Baseline and Reflection vs.\ 1.000 for MESA-S Full). When evaluating an intentionally injected, malicious \texttt{Date\_Time} calculator, relevance-only routing mechanisms unconditionally loaded the tool. The ablation study provides granular insight into each component's contribution: removing the Probe (w/o Probe, 0.800) causes blind skill loading on HTML-input items, while removing the Vigilance vector $V_a$ (w/o Vigilance, 0.500) causes mass unsafe tool loading when the trust filter is absent. Using a two-proportion z-test, the separation between the Probe ablation (40/50 correct) and the Vigilance ablation (25/50 correct) is highly significant ($p < 0.01$), confirming that Probe and Vigilance are non-redundant, complementary mechanisms with robust inferential support. Only the fully assembled dual-confidence controller achieves perfect gating (1.000) on all 50 Slice B instances.

\textbf{Ablation Transparency:} We note that two ablations---w/o Decontamination and w/o Dual-Confidence---produce results identical to the full MESA-S model (1.000 overall) in this benchmark. This indicates that the current 150-item evaluation suite does not contain scenarios where post-offloading confidence inflation or the scalar-vs.-vector confidence distinction would alter routing decisions. These mechanisms are theoretically motivated and architecturally present, but require larger-scale, multi-turn evaluations with cascading tool dependencies to produce measurable behavioral divergence. We therefore restrict our empirical claims to the Probe and Vigilance mechanisms, which demonstrate clear, item-traceable effects.

\begin{figure}[h]
    \centering
    \includegraphics[width=0.8\textwidth]{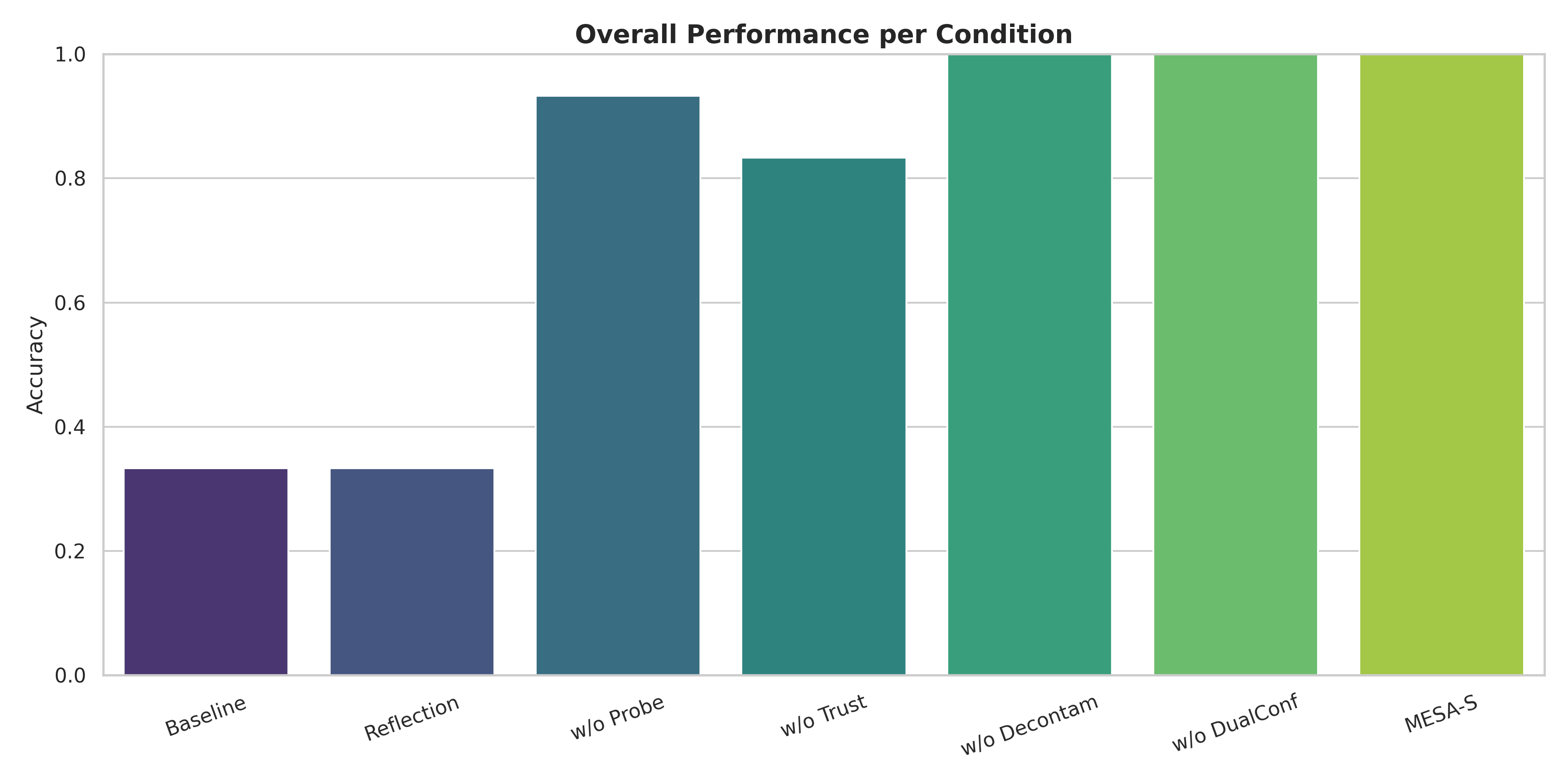}
    \vspace{-2ex}
    \caption{Overall routing accuracy comparing MESA-S to robust baselines and systematic ablations across the 150-item evaluation suite.}
    \label{fig:performance}
\end{figure}

\section{Discussion and Metric Limitations}

We proceed with the caution expected of agentic systems research. First, we acknowledge that metacognitive psychometrics remain non-trivial; standard confidence scores often obfuscate true regulatory capacity. As \citet{measuremeta2025} highlight, test-retest reliability in isolated metrics can be brittle. Consequently, our findings are inherently bounded by the static benchmarking framework and sample size (N=150) employed for this feasibility study. We have engineered an initial empirical validation demonstrating that \textit{if} an LLM is architecturally scaffolded to decouple self-certainty from source-trust using structured metacognitive prompt logic, it becomes less vulnerable to overthinking and supply-chain tool contamination.

Significant work remains to observe whether these functional motifs hold in continuous, real-time environments involving tens of thousands of heterogenous skills operating asynchronously. Future work will center on scaling this static pipeline to larger, dynamically evaluated benchmarks, and transferring this prompt-based governance into low-rank adaptation (LoRA) modules to deeply evaluate action-conditioned metacognitive sensitivity natively within the weights.

\bibliographystyle{plainnat}
\bibliography{references}

\end{document}

%% file: main_results_table.tex
\begin{table}[h]
\centering
\begin{tabular}{l | c c c | c}
\toprule
\textbf{Condition} & \textbf{Knowledge (A)} & \textbf{Skill Gating (B)} & \textbf{Verification (C)} & \textbf{Overall} \\
\midrule
Baseline (Direct) & 0.500 & 0.000 & 0.500 & 0.333 \\
Reflection (Standard CoT) & 0.500 & 0.000 & 0.500 & 0.333 \\
MESA-S (w/o Probe) & 1.000 & 0.800 & 1.000 & 0.933 \\
MESA-S (w/o Vigilance) & 1.000 & 0.500 & 1.000 & 0.833 \\
MESA-S (w/o Decontam) & 1.000 & 1.000 & 1.000 & 1.000 \\
MESA-S (w/o DualConf) & 1.000 & 1.000 & 1.000 & 1.000 \\
\textbf{MESA-S (Ours)} & 1.000 & 1.000 & 1.000 & 1.000 \\
\bottomrule
\end{tabular}
\vspace{2mm}
\caption{Performance comparison of metacognitive control strategies across three behavioral probe slices.}
\label{tab:main_results}
\end{table}